\documentclass{bmvc2k}

\usepackage{CJKutf8}
\usepackage{graphicx}
\usepackage{floatrow}
\usepackage{makecell}
\usepackage{multirow}
\usepackage{amssymb}
\graphicspath{ {./images/} }
\newfloatcommand{capbtabbox}{table}[][\FBwidth]
\usepackage[normalem]{ulem}
\useunder{\uline}{\ul}{}
\usepackage{subcaption}

\usepackage{pifont}

\title{Stylistic Multi-Task Analysis of Ukiyo-e Woodblock Prints}

\addauthor{Selina Khan}{selinajasmin@gmail.com}{1}
\addauthor{Nanne van Noord}{n.j.e.vannoord@uva.nl}{1}

\addinstitution{
 Informatics Institute\\
 University of Amsterdam\\
 Amsterdam, NL
}

\runninghead{Khan, van Noord}{Stylistic Multi-task Analysis}

\def\eg{\emph{e.g}\bmvaOneDot}

\def\etal{\emph{et al}\bmvaOneDot}
\def\ie{\emph{i.e}\bmvaOneDot}

\begin{document}

\maketitle

\begin{abstract}
In this work we present a large-scale dataset of \textit{Ukiyo-e} woodblock prints. Unlike previous works and datasets in the artistic domain that primarily focus on western art, this paper explores this pre-modern Japanese art form with the aim of broadening the scope for stylistic analysis and to provide a benchmark to evaluate a variety of art focused Computer Vision approaches. Our dataset consists of over $175.000$ prints with corresponding metadata (\eg artist, era, and creation date) from the 17th century to present day.
By approaching stylistic analysis as a Multi-Task problem we aim to more efficiently utilize the available metadata, and learn more general representations of style.
We show results for well-known baselines and state-of-the-art multi-task learning frameworks to enable future comparison, and to encourage stylistic analysis on this artistic domain. 
\end{abstract}

\section{Introduction}
\label{sec:intro}
Stylistic analysis of artwork has received increased attention with the emergence of Deep Learning and the growing quantities of visual artistic data made available by heritage institutions \cite{Mensink2014, Khan2014, mao, carneiro}. Within the domain of art history such stylistic analysis is used in the study of formalism. This involves studying artworks by analyzing and comparing form and style \cite{bell1914art}. Here, ‘style’ refers to the resemblance artworks have to one another in terms of solely visual aspects. If many visual elements are shared by enough artworks, their combination is distinctive and recognizable. Interpreting such stylistic aspects helps art historians to identify the artworks’ corresponding artist, year of creation, art movement, or geographical origin. When approaching formalism from a Computer Vision perspective it becomes clear that this problem is inherently multi-task that requires inferring multiple metadata aspects based on visual aspects of the artwork. 

Approaching stylistic analysis as a multi-task problem aids in overcoming one of the biggest challenges of the artistic domain: the lack of data. For most artists only a small number of artworks are available, and although humans are very good at generalizing from a few examples - vision models remain limited in this respect. However, by leveraging the relations between tasks in a multi-task setting it becomes possible to learn more general representations \cite{Caruana_1997}. This was for instance shown by Strezoski \etal. \cite{strezoski2017omniart}, who reported state-of-the-art results on the Rijksmuseum Challenge \cite{Mensink2014} with significantly shorter training duration. The suitability of multi-task learning for artwork has led to a growing amount of publicly available context-rich artwork datasets \cite{strezoski2017omniart, Khan2014, 10.1145/3123266.3123405}. However, a large part within the artistic domain that is neglected for stylistic analyses are non-western artworks. Most large-scale art databases with metadata consist of mainly western and European art. As such in this in this paper we focus specifically on Japanese \textit{Ukiyo-e} woodblock prints.  

\begin{figure}[!t]
    \centering
    \raisebox{-0.15\height}{\includegraphics[width=0.2\linewidth]{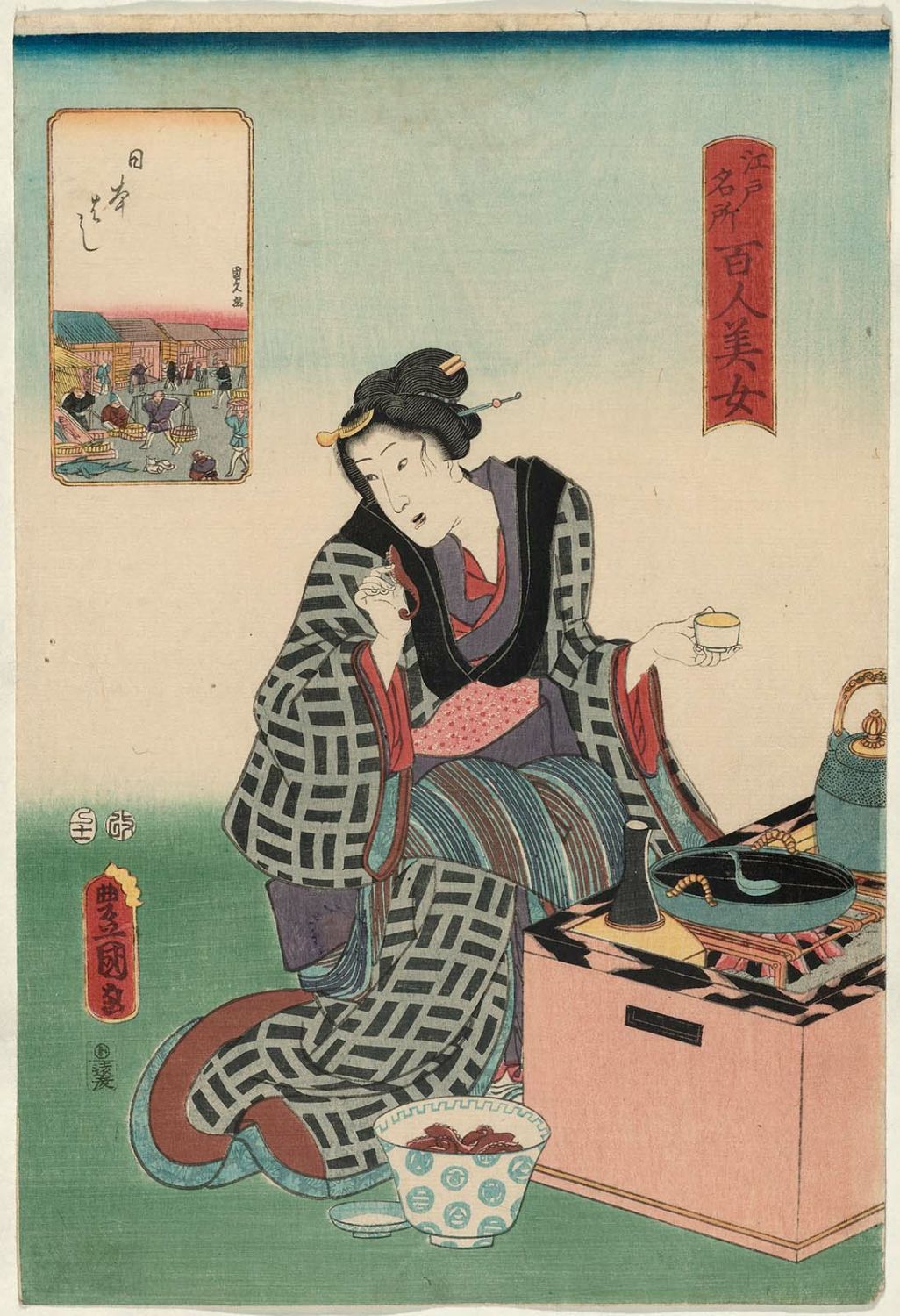}}
    \enskip
    \resizebox{5cm}{!}{%
    \begin{tabular}[b]{cc}\hline
      Attribute & Value \\ \hline \\
      Title & \begin{CJK}{UTF8}{min}「江戸名所百人美女」 「日本はし」\end{CJK} \\
            &  One Hundred Beautiful Women \\ & at Famous Places in Edo : Nihombashi. \\
      Artist & Utagawa Kunisada \\
      Date & 1857 \\
      Era & Popularization of Woodblock Printing \\
      Description &  \begin{CJK}{UTF8}{min}一般/江戸/名所案内記、図絵/\end{CJK}\\ 
            &  General - Edo's Famous Places. \\
      Source & Waseda University Theatre Museum \\ \\ \hline
    \end{tabular}}
    \caption{Ukiyo-e print with attributes from the proposed dataset.}
    \label{fig:metadata}
\end{figure}

\textit{Ukiyo-e} \begin{CJK}{UTF8}{min}(浮世絵)\end{CJK} is a genre of pre-modern Japanese artworks which flourished from the 16th through the 19th century. The term translates as “pictures of the floating world” which refers to the hedonic spirit of lower classes at the time. This spirit is depicted in daily subjects, such as kabuki theaters, geishas, landscapes, animals, and plants. These artworks usually took the form of either woodblock prints or paintings. Ukiyo-e woodblock prints are of particular interest for stylistic analysis, since identifying such prints can be a difficult task. Both the dyes and paper used for such prints are sensitive to light and seasonal changes, causing the prints to fade over time. Since Ukiyo-e prints were mass-produced, multiple versions of the same print are quite common. This leaves us with a vast collection of artworks in this genre, but it also results in significant variation in terms of condition, rarity, and quality between identical prints \cite{harris2011ukiyo}. For instance, some prints may have stains, creases, tears, or the prints may have been retouched later on. Additionally, woodblock carvers may have changed the colors or composition of prints that went through multiple editions over time, and the paper may have been trimmed within different margins \cite{hillier_1960}. 

In a number of recent works, \textit{Ukiyo-e} has been investigated on a relatively small scale, primarily by using feature engineering approaches \cite{tian2020kaokore, tian2021ukiyoe}. To make it possible to explore state-of-the-art models in this domain, and to push forward the state of stylistic analysis for a wider variety of art we make the following contributions: 1) We present a challenging new large-scale dataset of over $175.000$ Ukiyo-e woodblock print images. 2) We formulate a new multi-task problem for this dataset, making use of the metadata paired with each image. 3) We evaluate a number of baseline models on this new dataset, providing a solid foundation for further experimentation and comparison, and demonstrating the challenges and potential for stylistic analysis of \textit{Ukiyo-e} artworks.

\section{Related Work}
Despite the increasing attention to artistic analysis, the data used for this has been predominantly European \cite{strezoski2017omniart, Mensink2014, Khan2014, DBLP:journals/corr/KarayevHWAD13}. Although large scale datasets such as Omniart \cite{strezoski2017omniart} do contain artworks from other regions, they are still heavily dominated by western art. 
Nonetheless, a few works have begun exploring \textit{Ukiyo-e} artworks through the construction of small-scale datasets of Ukiyo-e artworks and faces \cite{tian2020kaokore, weko_3982_1}, for instance for use in a generative settings \cite{VERNIER2020100026, pinkney2020ukiyoe}.
Additionally, Tian et al. \cite{tian2021ukiyoe} presented an \textit{Ukiyo-e} dataset of $11,000$ annotated paintings and faces and used this to perform a quantitative study of faces in Ukiyo-e artworks. Using engineered geometry features they demonstrated the potential for automatically distinguishing Ukiyo-e styles. In this paper we present a large-scale dataset of over $175,000$ Ukiyo-e woodblock prints that makes it possible to train Deep Learning models. We use this dataset to build on this prior work to further the artistic analysis of Ukiyo-e art whilst incorporating a multi-task approach.

Although our proposed dataset is significantly bigger than previous datasets for \textit{Ukiyo-e} art, it can still be considered small-scale when compared to the datasets used to train state-of-the-art vision models \cite{dosovitskiy2020image}. However, \textit{Transfer learning} has proven to be a successful method for solving this particular problem, as it makes it possible to fine-tune models, originally trained on large-scale datasets of natural images, for the artistic domain \cite{Yosinski_Clune_Bengio_Lipson_2014,Milani_Fraternali_2020}. Examples of the successes of transfer learning for artwork analysis using CNN’s include people recognition \cite{Westlake2016}, object detection and labelling \cite{crowley2016art,7472087, Crowley14}, and determining genre and style in artworks \cite{7123719, CETINIC2018107}. 
Banerji et al. \cite{Banerji2017} experiment with transfer learning using pre-trained models and training models from scratch, similarly to Tan et al. \cite{7533051}, confirming the benefits of Transfer Learning for the artistic domain. In addition, Milani and Fraternali \cite{Milani_Fraternali_2020} recently showed that transfer learning in the artistic domain is highly effective when only fine-tuning the later layers of a network. In our work we follow this approach to transfer learning, and compare it to using pre-trained networks as feature extractors. 

A key aspect of stylistic analysis is relating known metadata quantities to visual components. In order to do this for models which are function as black boxes, we need to rely on interpretability methods \cite{9050545}. For CNN's, the model which is primarily used in existing stylistic analysis literature, this requires additional steps with methods that are applied post-hoc \cite{DBLP:journals/corr/ZeilerF13, simonyan2014deep}. Recent developments have shown that Vision Transformers (ViT) are not only a great addition to the field in terms of performance \cite{dosovitskiy2020image}, but they also offer a lot of potential for interpretability \cite{DBLP:journals/corr/abs-2012-14214}. Unlike CNN-based interpretability methods, which depend on a separate analysis using gradient-based methods, ViT models make use of attention to arrive at predictions. Because of the spatial nature of this attention, \ie, ViT can attend differently to different spatial areas, the attention mechanism of ViT can be directly used to create heat map visualisations \cite{DBLP:journals/corr/abs-2005-12872}. In this paper we use this capability of ViT to explore the predictions made for a variety of Ukiyo-e prints, combined with further qualitative analysis of the results, to get insights into the stylistic analysis. 

\begin{figure}[!ht]
\centering
\begin{subfigure}[t]{0.49\textwidth}
\centering
\includegraphics[width=40mm]{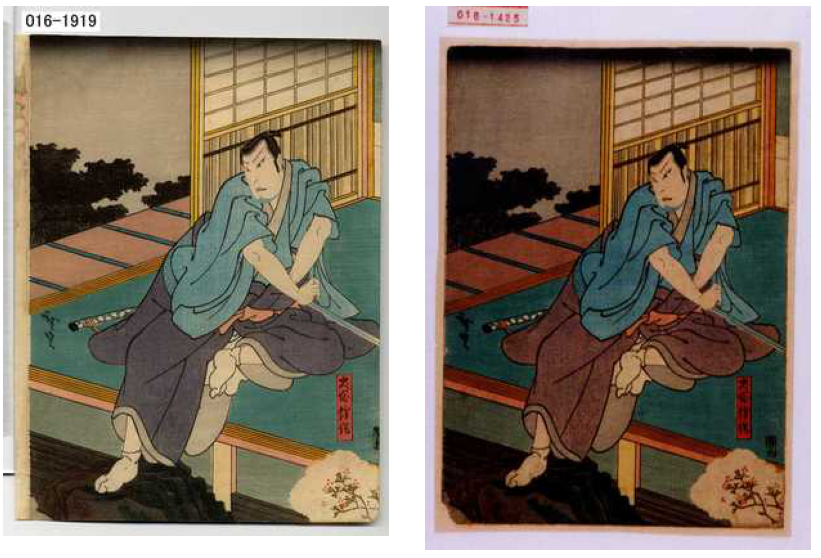}
\caption{Multiple editions of the same print.}
\label{fig:a}
\end{subfigure}
\begin{subfigure}[t]{0.49\textwidth}
\centering
\includegraphics[width=60mm]{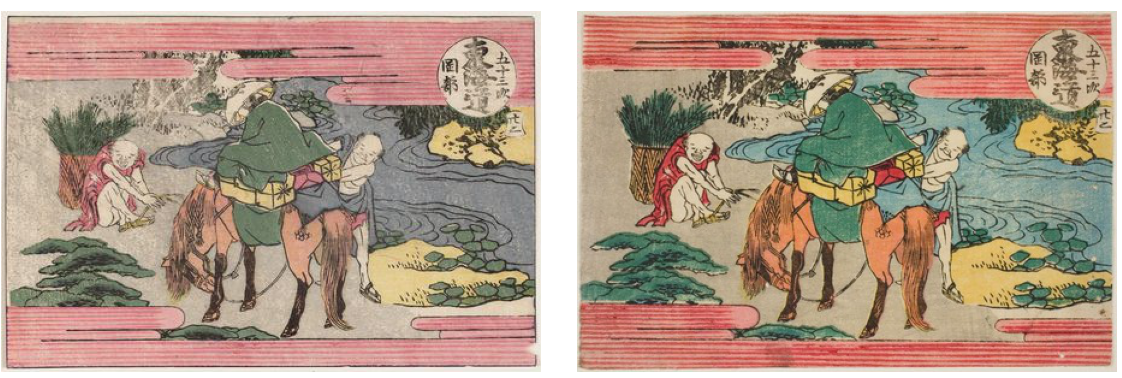}
\caption{Color differences between the same prints.}
\label{fig:b}
\end{subfigure}
\caption{Examples of pairs of Ukiyo-e prints made from the same woodblock but with a different appearance.}
\end{figure}

\section{Ukiyo-e dataset}
For the dataset presented in this research were build on the ukiyo-e.org database.\footnote{\url{https://ukiyo-e.org/}} The ukiyo-e.org database is a website with over $220,000$ Japanese woodblock prints dating from the early 17th century to present day. These images on the website are aggregated from a variety of museums, universities, libraries, auction houses, and dealers around the world. The ukiyo-e.org website makes it possible to search for Japanese woodblock prints and to compare similar prints across multiple collections. Additional examples images from our Ukiyo-e dataset can be found in Appendix~A. 

To construct the \textit{Ukiyo-e} dataset\footnote{The dataset and accompanying evaluation code will be publicly available on \url{https://github.com/selinakhan/stylistic-MTL-ukiyoe}} we collected all images and associated metadata from the ukiyo-e.org website, which after the removal of broken images resulted in a dataset of $177,897$ images. Among these images are a lot of duplicate artworks. These images are not exact copies; they are physically different artworks, but resemble the same image. Such duplicates include reprints~\ref{fig:a}, later editions using re-carved blocks, or (colour-faded) copies from the same woodblock~\ref{fig:b}. To gain more insights into the dataset, the cosine similarity between every image was used to find duplicates. If the cosine similarity between two images is $>94\%$, it is classified as a duplicate. Therefore, to enrich the dataset's metadata, each image is also labelled with each image it is highly similar to. 

It is conventional to split the dataset into a training, validation and test set. For our experiments, the dataset was split to use 80\% of the dataset for training, and 10\% for both the validation and test set. Since the dataset is rather large, we do not need larger testing or validation sets, because ten percent of the entire dataset is already nearly 18.000 images. 

Although there are number of available attributes per image (see Figure~\ref{fig:metadata}), we focus on the artwork's artist, date and era in our research. Upon inspection of the dataset, we found that the distribution of classes is rather skewed. The \textbf{artist} attribute is annotated for every image in the dataset, with the top-5 producing artist dominating 50\% of the entire dataset and 149 artists in total. Four of these artists were active within the same era, which is reflected in the distribution of the \textbf{era} attribute. Each artwork in the dataset is labelled with one of the seven era's, where 53\% of all artworks fall withing the era \textit{Popularization of Woodblock Printing (1804 to 1868)}. For roughly $25$\% of the artwork the exact creation \textbf{date} is not known or too vaguely specified (\eg, ``18th century''). For our following experiments, these dates are generated by uniformly sampling a date within the artwork's creation era. 

\section{Stylistic Multi-Task Artwork Analysis}

The aim of stylistic artwork analysis is to link visual elements to certain known quantities, for instance, when dealing with artist attribution the goal is to learn which visual elements enable us to recognise an artist's work \cite{Johnson_Hendriks_Berezhnoy_Brevdo_Hughes_Daubechies_Li_Postma_Wang_2008}. By linking these visual elements to the artist we gain insight into the artist's \text{style}. However, an artist does not produce their art in isolation, they are influenced by those that came before them, by contemporary artists, and also by techniques and materials that were at that time available. Art historians take this context into account when analysing artworks, and as such we argue that automatic stylistic analysis should do the same. Hence, we propose to model the stylistic analysis as a multi-task problem, predicting not only the artist of an artwork, but also when it was made, in terms of era and in terms of the creation year.

\begin{figure}[!h]
    \centering
    \includegraphics[width=0.8\linewidth]{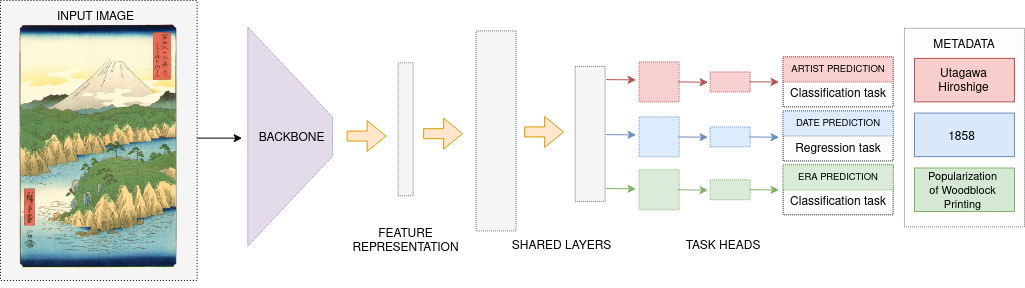}
    \caption{Stylistic Multi-task analysis model architecture.}
    \label{fig:mtl}
\end{figure}

In terms of modelling decisions there are a variety of architectural components that influence the performance of an stylistic analysis model. To assess the influence of these components on the task of determining an artwork's artist, era and date, we compare a number of variants of the training pipeline. We will first experiment with each task in a single-task learning setting. These results will be used as a baseline to evaluate the performance of the multi-task learning approach, and will highlight how different tasks perform paired with each other task. For the multi-task models we will use a well known and widely used baseline, that trains multiple task heads on top of a single shared backbone \cite{Caruana_1997,strezoski2017omniart}, as illustrated in Figure~\ref{fig:mtl}
Besides conducting experiments in a single- and multi-task learning setting, the backbones used in the experiments are also altered. A ResNet50 \cite{he2015deep} and ViT backbone will be incorporated to compare the performance of using CNN's and ViT's for a computational stylistic analysis.

Additionally, we perform an exploration of transfer learning methods. Although our proposed dataset is much larger than existing Ukiyo-e datasets, it cannot be used to adequately train state-of-the-art models from scratch. However, as transfer learning has shown great promise in the artistic domain we instead turn to this and aim to find the appropriate configuration here \cite{Milani_Fraternali_2020}. We experiment with two configurations: in the first setting the entire model will be kept frozen, this effectively translates to training a classifier on top of a fixed and pre-trained feature extractor. In the second setting only part of the model is frozen, by only fixing the parameters of the earlier layers of the backbone. This will result in the image's high-level features to be more specifically fine-tuned on artistic data. This method also provides additional insights when visualising attention maps, because the backbone is trained on the artwork images. Visualising the regions to which the model attends can be especially meaningfully when comparing a single- and multi-task results. This second transfer learning approach has previously been shown to outperform more complex transfer learning approaches \cite{Yosinski_Clune_Bengio_Lipson_2014}, this result has recently been reproduced in the artistic domain \cite{Milani_Fraternali_2020}.

For both multi-task and transfer learning there are methods that tackle these problems in more advanced manners, making better use of the available data and relationships between tasks \cite{Strezoski_Noord_Worring_2019,Vandenhende_Georgoulis_Van_Gool_2020,cross-stitch2016,nddr2019,mtan2019}. Although our aim with this work is not to provide an exhaustive comparison of all possible methods and models, we do perform an additional experiment to compare the performance of models trained with \textit{hard sharing} (i.e., sharing all model weights apart from the task heads) to three well-established MTL frameworks: Cross-Stitch Networks \cite{cross-stitch2016}, NDDR-CNN \cite{nddr2019}, and MTAN \cite{mtan2019}. These MTL frameworks optimize the manner in which information is shared between tasks throughout the network, with the aim of improving MTL performance. Through these evaluations we hope to establish a clear and reliable benchmark to facilitate future comparison. 

\section{Experimental Setup}
We perform two sets of experiments. In the first we introduce a range of baselines in a variety of settings to demonstrate the influence of various modeling decisions. In the second set of experiments we evaluate the performance of existing MTL frameworks on our dataset, and compare them to our baselines. We focus on two backbones to obtain baselines on our proposed Ukiyo-e dataset, a ViT\footnote{Using the implementation at: \url{https://github.com/jeonsworld/ViT-pytorch}} \cite{dosovitskiy2020image} and a ResNet \cite{He_Zhang_Ren_Sun_2016} model. The ViT model is pre-trained on the ImageNet-21K dataset \cite{DBLP:journals/corr/abs-2104-10972}, whereas the ResNet is pre-trained on the smaller and more widely used 1M images ImageNet split \cite{Russakovsky_Deng_Su_Krause_Satheesh_Ma_Huang_Karpathy_Khosla_Bernstein_2015}. For the ViT we use the ViT-Base\_16\footnote{ViT-Base\_16 checkpoint: \url{https://console.cloud.google.com/storage/browser/vit_models;tab=objects?prefix=&forceOnObjectsSortingFiltering=false}} configuration, with an image grid of $16 \times 16$, $12$ layers, a hidden dimension size of $768$, and MLP size of $3072$, and $12$ heads. For the ResNet we use the 50 layer configuration, \ie, ResNet50, from the Torchvision library.

For each of the two backbones we explore two approaches to Transfer Learning: 1) freezing the entire backbone, or 2) partially freezing the backbone. Freezing the backbones is much more efficient, as the backbone is essentially just used a feature extractor and it does not need to be optimized. However in turn, the features in the backbone cannot be adjusted to the specific task(s) and domain when frozen. The partially frozen backbones offers a good compromise, as it makes it possible to fine-tune the features, while still being relatively efficient. We implement the partial freezing by disabling the gradient computation during forward passes for the first half of the backbone. Freezing only the earlier layers of the backbone preserves the lower-level primitive features learned, and optimizes the more detailed high-level features to fit images from the artistic domain in the higher layers. For all models we use early-stopping on the validation set to obtain the final checkpoint to evaluate on the test set.

\subsection{Single-task learning}
The single- and multi-task model configurations differ in the number of task heads they optimize. The single task models consist of exactly one task head that optimizes its fully-connected layers for the specific task at hand. It uses a different backbone, depending on the experiment. The model's backbone is followed by three fully connected layers where the final layer outputs a prediction. To improve the performance of the task head, a rectified linear activation unit (ReLU) is placed between every fully connected layer. The ReLU function ensures that the output of the layers remain positive by setting every negative output to zero. This helps to overcome the vanishing gradient problem, allowing the model to learn faster and perform better \cite{Goodfellow-et-al-2016}. For the classification tasks, a cross-entropy loss function is minimized during training. The regression task is minimized using the L1 loss. A single-task model is trained for each individual task, experimenting with a ResNet50, frozen ViT, and partially frozen ViT as a backbone.

\subsection{Multi-task learning}
The multi-task model (see Figure~\ref{fig:mtl}) consists of $n$ task heads, where $n$ is the number of tasks to be trained simultaneously. In Figure~\ref{fig:mtl}, the multi task model trains three tasks simultaneously. Akin to the single-task model, the multi-task model has an interchangeable backbone, and has a ReLU function placed in between each layer. The backbone is followed by two shared fully-connected layers, enabling the model to learn relations between tasks. Finally, the model splits up in $n$ task heads consisting of two fully-connected layers before making a task-specific prediction, this type of multi-task learning is commonly referred to as \textit{hard sharing} \cite{strezoski2017omniart, sener_neurips2018, Strezoski_Noord_Worring_2019, pmlr-v119-guo20e}. 
Training a model in a multi-task setting involves jointly optimizing more than one loss function \cite{Caruana_1997}. To avoid problems of an imbalanced loss due to the combination of losses that fall within a different range \cite{Cipolla_Gal_Kendall_2018}, the L1 date loss is scaled down to fall within the range of the classification losses (\ie, artist and era). Where the cross-entropy loss function outputs a number between 0 and 1, the L1 loss outputs a number within a much larger range. Scaling the L1 loss ensures that it does not dominate the joint loss, due to its range being significantly larger. The scalar $\sigma$ for the regression loss was optimized on the validation set and a value of $1000$ was found to be optimal. For all multi-task models, the sum of the (scaled) losses of each task is optimized. The loss function when training the task of predicting the artist, date and era simultaneously is shown in Equation~\ref{eq:loss}.

\begin{equation} \label{eq:loss}
L_{MTL} = CrossEntropyLoss(artist) + \frac{L1Loss(date)}{\sigma} + CrossEntropyLoss(era)
\end{equation}

For the multi-task setting we compare all pairwise combinations of the three tasks (\ie. artist + era, artist + date, and date + era), additionally we also evaluate training all three tasks jointly.

\subsection{MTL Frameworks}

We additionally evaluate three state-of-the-art MTL frameworks: Cross-Stitch Networks \cite{cross-stitch2016}, NDDR-CNN \cite{nddr2019}, and MTAN \cite{mtan2019}. Cross-Stitch Networks combine the activations from multiple networks using cross-stitch units, these units learn linear combinations of the activations of the networks and share this combination with each network - distributing the task information. Conceptually, NDDR-CNN is similar to Cross-Stitch, except that it fusing at every layer. MTAN deviates from this concept by learning a single shared feature representation, which is then adapted to task-specific representations by means of attention layers. All three MTL frameworks were designed around CNN architectures, as such we implement them using a ResNet50 backbone.\footnote{For the MTL frameworks we use the excellent repo at: \url{https://github.com/SimonVandenhende/Multi-Task-Learning-PyTorch}}

\section{Results}
To assess the performance per specific task and model configuration, the accuracy score will be used in determining how well the model attributes unseen artworks. The accuracy measure directly reflects the classification performance, but the date regression task it cannot be directly used. Instead, we evaluate predicting the date of creation as if it were a classification task by using a threshold. For every predicted date, if the date is at most $20$ years off the original date, the prediction is classified as correct. Furthermore, we evaluate the date task with the Mean Absolute Error (MAE) as shown in table~\ref{table:results}

\begin{table}[h]
\scriptsize
\begin{tabular}{l|l|c|l|l|l|l}
Task(s) & Backbone & Frozen & Artist $\uparrow$ & Era  $\uparrow$ & Date $\uparrow$ & \shortstack{Date \\ (MAE) $\downarrow$ } \\\hline \hline
\multirow{4}{*}{Single-task} & \multirow{2}{*}{ResNet50} & \checkmark & $54.4$ &  $79.4$ & $37.1$ & $34.8$\\
                      & & $\times$ & $76.5$ &  $90.6$ & $68.8$ & $17.9$\\\cline{2-7}
& \multirow{2}{*}{ViT} & \checkmark & $56.5$ & $74.1$ & $50.1$  & $26.1$\\
                         &  & $\times$ & $\mathbf{82.1}$ & $92.1$ & $51.1$ & $25.8$\\\hline\hline
\multirow{4}{*}{Artist \& Era} & \multirow{2}{*}{ResNet50} & \checkmark & $30.7$ & $75.3$ & - & -\\
                          & & $\times$ & $41.8$ & $85.4$ & - & -\\ \cline{2-7}
& \multirow{2}{*}{ViT} & \checkmark & $66.5$ & $86.7$ & - & -\\ 
                          & & $\times$ & $72.8$ & $\mathbf{92.7}$ & - & -\\\hline
\multirow{4}{*}{Artist \& Date} & \multirow{2}{*}{ResNet50} & \checkmark & $27.4$ & - & $23.4$ & $55.7$\\
                          & & $\times$ & $34.6$ & - & $44.3$ & $29.1$\\ \cline{2-7}
& \multirow{2}{*}{ViT} & \checkmark & $18.1$ & - & $44.9$ & $24.0$ \\ 
                          & & $\times$ & $37.8$ & - & $56.8$ & $23.8$\\\hline
\multirow{4}{*}{Era \& Date} & \multirow{2}{*}{ResNet50} & \checkmark & - & $76.9$ & $44.4$ & $29.4$ \\
                          & & $\times$ & - & $86.2$ & $48.6$ & $27.3$\\ \cline{2-7}
& \multirow{2}{*}{ViT} & \checkmark & - & $74.3$ & $52.6$ & $25.2$\\ 
                          & & $\times$ & - & $89.5$ & $\mathbf{71.4}$ & $\mathbf{16.6}$\\\hline
\multirow{4}{*}{Artist, Era \& Date} & \multirow{2}{*}{ResNet50} & \checkmark & $29.3$ & $74.7$ & $33.3$ & $39.1$ \\
                          & & $\times$ & $38.9$ & $85.4$ & $38.1$ & $31.3$\\ \cline{2-7}
& \multirow{2}{*}{ViT} & \checkmark & $30.3$ & $71.7$ & $52.4$ & $25.1$\\
                          & & $\times$ & $70.1$ & $92.2$ & $70.5$ & $16.9$\\\hline
\end{tabular}
\caption{Results on the Ukiyo-e dataset for the comparison between different backbones, transfer learning approaches, and combinations of tasks. Best score per task in bold.} 
\label{table:results}
\end{table}

Our results for the baselines in Table~\ref{table:results} shown that when comparing the frozen to the partly frozen architectures that the latter significantly outperforms the former, for both the ResNet \cite{He_Zhang_Ren_Sun_2016} and the ViT \cite{dosovitskiy2020image} architectures. This highlights that the visual appearance of the prints, and the features necessary to perform these tasks, strongly differ from ImageNet. Moreover, it shows the benefits of this Transfer Learning approach and that these benefits not only apply to CNN-based models as shown in previous work, but also apply to Vision Transformers. When comparing architectures, it is clear that the ViT backbone outperforms the ResNet50 backbone (with the exception of single task date prediction). A factor in this might be that the ViT model has more training parameters, particularly in the partly frozen setting, but we also see a clear interaction with the multi-task setup here. 

Across the tasks we see that artist and era prediction works particularly well, with date prediction trailing behind. In part, this is due to the noisy nature of date prediction, but similarities in style and re-use of woodblocks also makes this a very difficult task. Nonetheless, we see that the date predictions clearly improve when paired to another task. Surprisingly, for era we see only very minor gains when analysed in a multi-task setting, and for artist we actually see a drop in performance. At this point we are unsure why artist suffer from this negative interference, as the other two tasks do benefit from the multi-task setting.

Through qualitative analysis it becomes clear that all the well performing models are affected by similar confusions between artists. Besides confusions within the same era, most misclassifications are between artists that went to the same art school, or between pupils and masters, or even for artists that took on multiple names. There is a lot of resemblance in style between clusters of Japanese woodblock artists, as the art was mostly organized into schools and movements which specialized in certain styles. 


The results in Table~\ref{table:sota} show that all frameworks offer significant performance improvements over the ResNet50 baseline, matching the idea that more rigorous and deliberate sharing is beneficial to MTL. Interestingly, the gains of the MTL frameworks are much less pronounced when compared to the ViT model, and the ViT model is even more accurate on the \textit{Date} task. Among the frameworks MTAN \cite{mtan2019} performs best, outperforming the other two frameworks on every task. Overall, the results show that MTL frameworks  outperform hard sharing, and hint at a promising future for ViT-based MTL frameworks.

\begin{table}[h]
\scriptsize
\begin{tabular}{l|l|c|l|l|l|l}
Task(s) & Backbone & MTL Model & Artist $\uparrow$ & Era  $\uparrow$ & Date $\uparrow$ & \shortstack{Date \\ (MAE) $\downarrow$ } \\\hline \hline

\multirow{5}{*}{Artist, Era \& Date} & ResNet50 & \multirow{2}{*}{Hard Sharing} & $38.9$ & $85.4$ & $38.1$ & $31.3$ \\
 & ViT &  & $70.1$ & $92.2$ & $\mathbf{70.5}$ & $16.9$ \\ \cline{2-7}
& \multirow{3}{*}{ResNet50} & Cross-Stitch \cite{cross-stitch2016} & $75.7$ & $90.6$ & $52.4$ & $25.1$ \\
 &  & NDDR-CNN \cite{nddr2019} & $69.5$ & $87.8$ & $59.4$ & $21.3$ \\
 &  & MTAN  \cite{mtan2019}& $\mathbf{81.1}$ & $\mathbf{94.2}$ & $61.2$ & $\mathbf{19.1}$ \\ \hline

\end{tabular}
\caption{Results of comparison of baselines (using hard sharing) with state-of-the-art MTL frameworks. Best score per task in bold.} 
\label{table:sota}

\end{table}


\begin{table}[h]
\scriptsize
\begin{tabular}{l|l|l|l}
\hline
Prediction          & \multicolumn{1}{l|}{Actual} & \# & Analysis                                                                                                                                                             \\ \hline
Utagawa Kunisada    & Utagawa Kunisada II         & 62                    & Kunisada II inherited the name after marrying his master's daughter.                                                                                             \\ \hline
Morikawa Chikashige & Toyohara Kunichika          & 51                    & \begin{tabular}[c]{@{}l@{}}Little is known about Chikashige, but we know he was the pupil of \\ famous Meji-artist Toyohara Kunichika.\end{tabular}                  \\ \hline
Ishikawa Toyonobu   & Torii Kiyohiro              & 42                    & \begin{tabular}[c]{@{}l@{}}The artists were active in the same era, and primarily produced \\ \textit{yakusha-e} (actors) and \textit{bijin-ga} (beautiful women) prints.\end{tabular} \\ \hline
Utagawa Yoshitora   & Utagawa Yoshikazu           & 32                    & Both artists were students from Utagawa Kuniyoshi.                                                                                                                   \\ \hline
Ohara Koson         & Shoson Ohara                & 30                    & \begin{tabular}[c]{@{}l@{}}This artist went by three different titles: Ohara Hōson, \\ Ohara Shōson and Ohara Koson.\end{tabular}                                     \\ \hline
Kawase Hasui        & Tsuchiya Koitsu             & 28                    & \begin{tabular}[c]{@{}l@{}}Both artist were top-producing Shin-Hanga artists, both \\ specialized in landscapes.\end{tabular}                                        \\ \hline
Takahashi Hiroaki   & Takahashi Shōtei            & 28                    & \begin{tabular}[c]{@{}l@{}}As a young artist, Hiroaki Takahashi was given the artistic\\ name Shōtei by his tutor.\end{tabular}                                      \\ \hline
Utagawa Hiroshige   & Utagawa Hiroshige II        & 25                    & \begin{tabular}[c]{@{}l@{}}Suzuki Chinpei inherited the name Hiroshige II following the death \\ of his master Hiroshige, whose daughter he married.\end{tabular}    \\ \hline
Katsukawa Shunko    & Katsukawa Shunsho           & 24                    & Both artist are students from the Katsukawa school.                                                                                                                  \\ \hline
\end{tabular}
\caption{Artists between whom there is the most confusion, ranked by misclassification frequency, with an analysis of the possible reason for the confusion.}
\label{table:misclas}
\end{table}

In table \ref{table:skewness}, we compare the accuracy score for both the top-5 producing artists and the remaining 145 artists. We observe an interesting difference between the two models, as the ViT backbones shows less difference in performance between the over- and underrepresented classes compared to the ResNet50 backbone. This highlights the ViT's capability to deal with class imbalance.

\begin{table}[h]
\scriptsize
\begin{tabular}{l|l|c|l|l}
Task & Backbone & Frozen & Top-5 Artists & Remaining 145 Artists\\\hline \hline
\multirow{4}{*}{Single-task Artist} & \multirow{2}{*}{ResNet50} & \checkmark & $60.9$ &  $39.5$ \\
                      & & $\times$ & $91.0$ &  $62.7$ \\\cline{2-5}
& \multirow{2}{*}{ViT} & \checkmark & $63.0$ & $49.8$\\
                         &  & $\times$ & $90.0$ & $74.3$ \\\hline
\end{tabular}
\caption{Accuracy for top 5 most frequent artists as compared to the remaining 145 artists on single-task artist classification.}
\label{table:skewness}
\end{table}

Misclassification within the Utagawa school are most common (see Table~\ref{table:misclas} and Appendix~B) reflecting the fact that it was the largest \textit{Ukiyo-e} school at the time. It was a Japanese custom for successful apprentices to take the names of their masters, which is also apparent in the most common misclassifications. It is remarkable that a model is able to distinguish between these highly related artists, as it shows that it does in fact learn to link visual elements to artist's style. Similarly for the era attribute nearly all misclassifications between era's are ones that precede or succeed one other. This aligns with artworks near era boundaries being similar, rather than there being an abrupt transition. 

\begin{figure}[!ht]
\centering
\begin{subfigure}[b]{0.18\textwidth}
\centering
\includegraphics[width=\textwidth]{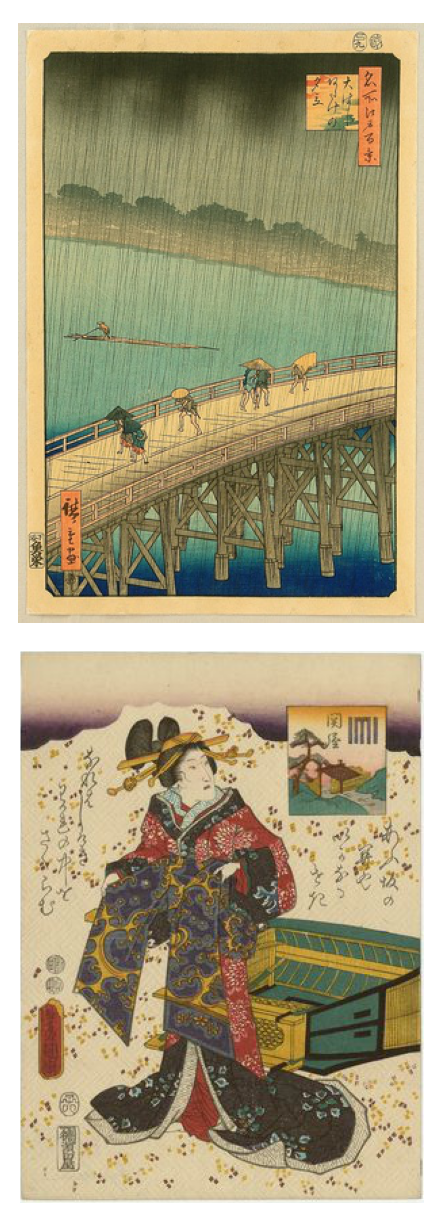}
\caption{Original}
\label{fig:att0}
\end{subfigure}
\begin{subfigure}[b]{0.18\textwidth}
\centering
\includegraphics[width=\textwidth]{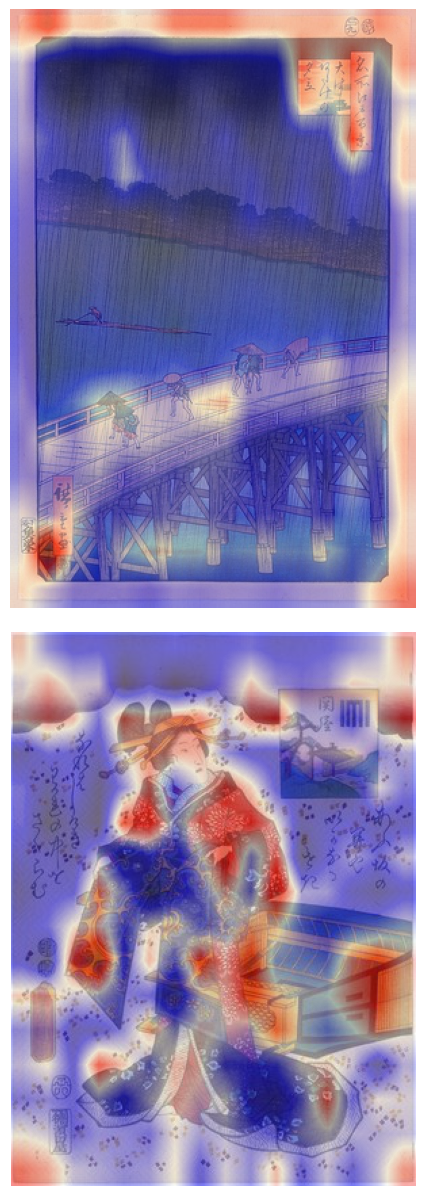}
\caption{}
\label{fig:att1}
\end{subfigure}
\begin{subfigure}[b]{0.18\textwidth}
\centering
\includegraphics[width=\textwidth]{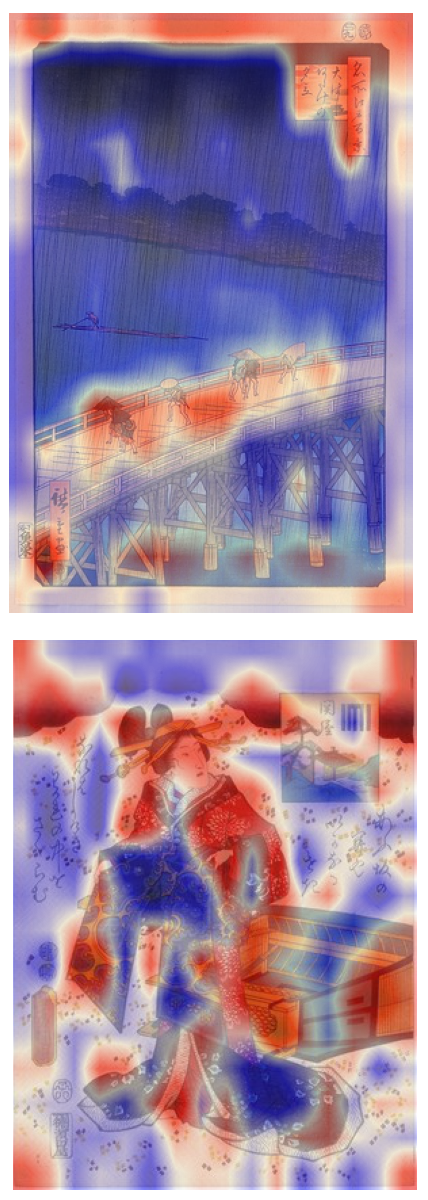}
\caption{}
\label{fig:att2}
\end{subfigure}
\begin{subfigure}[b]{0.18\textwidth}
\centering
\includegraphics[width=\textwidth]{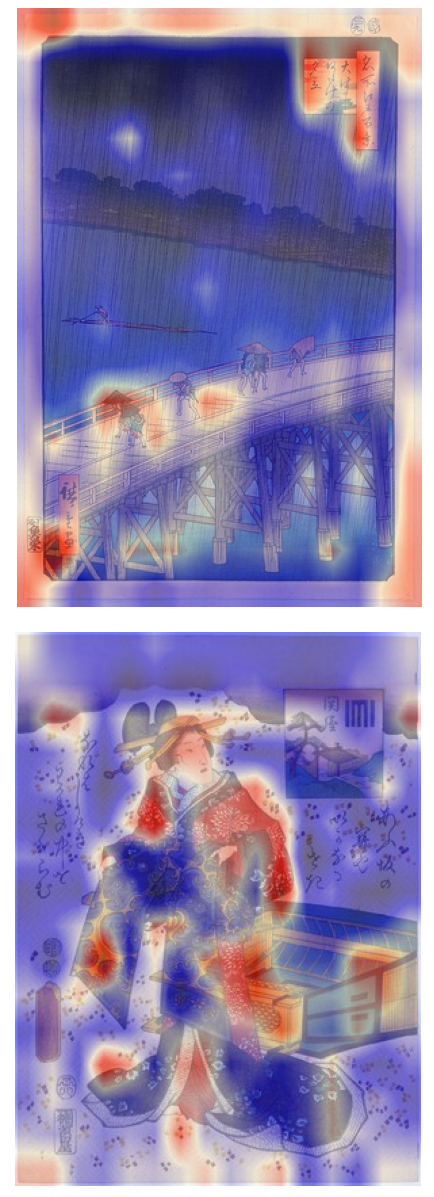}
\caption{}
\label{fig:att3}
\end{subfigure}
\begin{subfigure}[t]{0.04\textwidth}
\centering
\raisebox{0.25\height}{\includegraphics[scale=0.25]{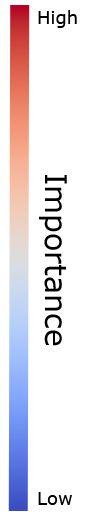}}
\label{fig:bar}
\end{subfigure}
\vspace*{-2mm}
\caption{Attention maps for partially frozen ViT model trained with single task artist prediction (b), single task era prediction (c), and multi-task artist \& era prediction (d).}
\label{fig:attmaps}
\end{figure}

Figure~\ref{fig:attmaps} shows the attention heat maps for the partially frozen ViT model. It is noticeable in the top row, for the print by Hiroshige, that the model focuses on his signature, in the upper right corner and bottom left corner. This indicates that the model makes use of Hiroshige's signature to recognize his artworks. However, for era prediction (c) and in the multi-task setting (d) we see that the model also attends to other parts of the print, focusing more on the content. In the bottom row we see that the model strongly attends to the person and her gown, and the horizon in the background. From these visualisations we can see that the model does attend to stylistic components in the image, but might also rely on other signals (\ie, a signature position) when possible.

\section{Conclusion}
In this work we presented a large-scale dataset of over $175,000$ \textit{Ukiyo-e} woodblock prints along with their metadata. To support stylistic analysis of Ukiyo-e prints we formulated the task of predicting an artwork's artist, era, and date in a multi-task learning setting, to leverage the relations between these tasks and learn a more general representation of style. Our results show that a multi-task learning approach benefits from the vision transformer backbone, with increases in performance up to $30\%$ compared to a ResNet backbone. In addition, the transformer model benefits more from our transfer learning strategy, as compared to the ResNet, and using an MTL framework the performance of a ResNet can be boosted to outperform a transformer model. This opens the door for ViT-based MTL frameworks for further performance gains.
Moreover, based on the stylistic analysis we are able to explain a variety of the misclassifications as confusions between students of the same master, or in some cases find artworks of an artist who went by different names. By releasing this dataset and presenting baselines for stylistic multi-task analysis using state-of-the-art deep learning models we hope to encourage future work on this understudied artistic domain.

\bibliography{egbib}

\begin{thebibliography}{46}
\providecommand{\natexlab}[1]{#1}
\providecommand{\url}[1]{\texttt{#1}}
\expandafter\ifx\csname urlstyle\endcsname\relax
  \providecommand{\doi}[1]{doi: #1}\else
  \providecommand{\doi}{doi: \begingroup \urlstyle{rm}\Url}\fi

\bibitem[wek()]{weko_3982_1}
Ritsumeikan ukiyo-e database.
\newblock text (tsv).
\newblock URL \url{https://doi.org/10.32130/rdata.2.1}.

\bibitem[Banerji and Sinha(2017)]{Banerji2017}
Sugata Banerji and Atreyee Sinha.
\newblock Painting classification using a pre-trained convolutional neural
  network.
\newblock In \emph{Computer Vision, Graphics, and Image Processing}, pages
  168--179. Springer International Publishing, 2017.
\newblock \doi{10.1007/978-3-319-68124-5_15}.
\newblock URL \url{https://doi.org/10.1007/978-3-319-68124-5_15}.

\bibitem[Bell(1914)]{bell1914art}
C.~Bell.
\newblock \emph{Art}.
\newblock Stokes, 1914.
\newblock URL \url{https://books.google.nl/books?id=-rgQAQAAMAAJ}.

\bibitem[Carion et~al.(2020)Carion, Massa, Synnaeve, Usunier, Kirillov, and
  Zagoruyko]{DBLP:journals/corr/abs-2005-12872}
Nicolas Carion, Francisco Massa, Gabriel Synnaeve, Nicolas Usunier, Alexander
  Kirillov, and Sergey Zagoruyko.
\newblock End-to-end object detection with transformers.
\newblock \emph{CoRR}, abs/2005.12872, 2020.
\newblock URL \url{https://arxiv.org/abs/2005.12872}.

\bibitem[Carneiro et~al.(2012)Carneiro, Silva, Del~Bue, and Costeira]{carneiro}
Gustavo Carneiro, Nuno Silva, Alessio Del~Bue, and Joao Costeira.
\newblock Artistic image classification: an analysis on the printart database.
\newblock pages 141--155, 10 2012.
\newblock ISBN 978-3-642-33764-2.
\newblock \doi{10.1007/978-3-642-33765-9_11}.

\bibitem[Caruana(1997)]{Caruana_1997}
Rich Caruana.
\newblock Multitask learning.
\newblock \emph{Machine learning}, 28\penalty0 (1):\penalty0 41–75, 1997.

\bibitem[Cetinic et~al.(2018)Cetinic, Lipic, and Grgic]{CETINIC2018107}
Eva Cetinic, Tomislav Lipic, and Sonja Grgic.
\newblock Fine-tuning convolutional neural networks for fine art
  classification.
\newblock \emph{Expert Systems with Applications}, 114:\penalty0 107--118,
  2018.
\newblock ISSN 0957-4174.
\newblock \doi{https://doi.org/10.1016/j.eswa.2018.07.026}.
\newblock URL
  \url{https://www.sciencedirect.com/science/article/pii/S0957417418304421}.

\bibitem[Cipolla et~al.(2018)Cipolla, Gal, and
  Kendall]{Cipolla_Gal_Kendall_2018}
Roberto Cipolla, Yarin Gal, and Alex Kendall.
\newblock Multi-task learning using uncertainty to weigh losses for scene
  geometry and semantics.
\newblock page 7482–7491. IEEE, Jun 2018.
\newblock ISBN 978-1-5386-6420-9.
\newblock \doi{10.1109/CVPR.2018.00781}.
\newblock URL \url{https://ieeexplore.ieee.org/document/8578879/}.

\bibitem[Crowley and Zisserman(2014)]{Crowley14}
Elliot~J. Crowley and Andrew Zisserman.
\newblock The state of the art: Object retrieval in paintings using
  discriminative regions.
\newblock In \emph{British Machine Vision Conference}, 2014.

\bibitem[Crowley and Zisserman(2016)]{crowley2016art}
Elliot~J Crowley and Andrew Zisserman.
\newblock The art of detection.
\newblock In \emph{ECCV Workshops (1)}, 2016.

\bibitem[Dosovitskiy et~al.(2020)Dosovitskiy, Beyer, Kolesnikov, Weissenborn,
  Zhai, Unterthiner, Dehghani, Minderer, Heigold, Gelly, Uszkoreit, and
  Houlsby]{dosovitskiy2020image}
Alexey Dosovitskiy, Lucas Beyer, Alexander Kolesnikov, Dirk Weissenborn,
  Xiaohua Zhai, Thomas Unterthiner, Mostafa Dehghani, Matthias Minderer, Georg
  Heigold, Sylvain Gelly, Jakob Uszkoreit, and Neil Houlsby.
\newblock An image is worth 16x16 words: Transformers for image recognition at
  scale, 2020.

\bibitem[Gao et~al.(2019)Gao, Ma, Zhao, Liu, and Yuille]{nddr2019}
Yuan Gao, Jiayi Ma, Mingbo Zhao, Wei Liu, and Alan~L. Yuille.
\newblock {NDDR}-{CNN}: Layerwise feature fusing in multi-task cnns by neural
  discriminative dimensionality reduction.
\newblock In \emph{CVPR}, 2019.

\bibitem[Goodfellow et~al.(2016)Goodfellow, Bengio, and
  Courville]{Goodfellow-et-al-2016}
Ian Goodfellow, Yoshua Bengio, and Aaron Courville.
\newblock \emph{Deep Learning}.
\newblock MIT Press, 2016.
\newblock \url{http://www.deeplearningbook.org}.

\bibitem[Guo et~al.(2020)Guo, Lee, and Ulbricht]{pmlr-v119-guo20e}
Pengsheng Guo, Chen-Yu Lee, and Daniel Ulbricht.
\newblock Learning to branch for multi-task learning.
\newblock In Hal~Daumé III and Aarti Singh, editors, \emph{ICML}, volume 119
  of \emph{Proceedings of Machine Learning Research}, pages 3854--3863. PMLR,
  13--18 Jul 2020.

\bibitem[Harris(2011)]{harris2011ukiyo}
F.~Harris.
\newblock \emph{Ukiyo-e: The Art of the Japanese Print}.
\newblock Tuttle Publishing, 2011.
\newblock ISBN 9784805310984.
\newblock URL \url{https://books.google.nl/books?id=Bs3DbwAACAAJ}.

\bibitem[He et~al.(2015)He, Zhang, Ren, and Sun]{he2015deep}
Kaiming He, Xiangyu Zhang, Shaoqing Ren, and Jian Sun.
\newblock Deep residual learning for image recognition, 2015.

\bibitem[He et~al.(2016)He, Zhang, Ren, and Sun]{He_Zhang_Ren_Sun_2016}
Kaiming He, Xiangyu Zhang, Shaoqing Ren, and Jian Sun.
\newblock Deep residual learning for image recognition.
\newblock In \emph{2016 IEEE Conference on Computer Vision and Pattern
  Recognition (CVPR)}, page 770–778. IEEE, Jun 2016.
\newblock ISBN 978-1-4673-8851-1.
\newblock \doi{10.1109/CVPR.2016.90}.
\newblock URL \url{http://ieeexplore.ieee.org/document/7780459/}.

\bibitem[Hillier(1960)]{hillier_1960}
J.~Hillier.
\newblock Japanese prints from the early masters to the modern. by james a.
  michener. charles e. tuttle company: Rutland, vermont, and tokyo, japan. in
  the u.s. 15.
\newblock \emph{Journal of the Royal Asiatic Society}, 92\penalty0
  (1-2):\penalty0 65–66, 1960.
\newblock \doi{10.1017/S0035869X00118817}.

\bibitem[Johnson et~al.(2008)Johnson, Hendriks, Berezhnoy, Brevdo, Hughes,
  Daubechies, Li, Postma, and
  Wang]{Johnson_Hendriks_Berezhnoy_Brevdo_Hughes_Daubechies_Li_Postma_Wang_2008}
C.~Richard Johnson, Ella Hendriks, Igor~J. Berezhnoy, Eugene Brevdo, Shannon~M.
  Hughes, Ingrid Daubechies, Jia Li, Eric Postma, and James~Z. Wang.
\newblock Image processing for artist identification.
\newblock \emph{IEEE Signal Processing Magazine}, 25\penalty0 (4):\penalty0
  37–48, Jul 2008.
\newblock ISSN 1558-0792.
\newblock \doi{10.1109/MSP.2008.923513}.

\bibitem[Karayev et~al.(2013)Karayev, Hertzmann, Winnemoeller, Agarwala, and
  Darrell]{DBLP:journals/corr/KarayevHWAD13}
Sergey Karayev, Aaron Hertzmann, Holger Winnemoeller, Aseem Agarwala, and
  Trevor Darrell.
\newblock Recognizing image style.
\newblock \emph{CoRR}, abs/1311.3715, 2013.
\newblock URL \url{http://arxiv.org/abs/1311.3715}.

\bibitem[Khan et~al.(2014)Khan, Beigpour, van~de Weijer, and
  Felsberg]{Khan2014}
Fahad~Shahbaz Khan, Shida Beigpour, Joost van~de Weijer, and Michael Felsberg.
\newblock Painting-91: a large scale database for computational painting
  categorization.
\newblock \emph{Machine Vision and Applications}, 25\penalty0 (6):\penalty0
  1385--1397, June 2014.
\newblock \doi{10.1007/s00138-014-0621-6}.
\newblock URL \url{https://doi.org/10.1007/s00138-014-0621-6}.

\bibitem[Liu et~al.(2019)Liu, Johns, and Davison]{mtan2019}
Shikun Liu, Edward Johns, and Andrew~J Davison.
\newblock End-to-end multi-task learning with attention.
\newblock In \emph{Proceedings of the IEEE Conference on Computer Vision and
  Pattern Recognition}, pages 1871--1880, 2019.

\bibitem[Mao et~al.(2017{\natexlab{a}})Mao, Cheung, and
  She]{10.1145/3123266.3123405}
Hui Mao, Ming Cheung, and James She.
\newblock Deepart: Learning joint representations of visual arts.
\newblock In \emph{Proceedings of the 25th ACM International Conference on
  Multimedia}, MM '17, page 1183–1191, New York, NY, USA, 2017{\natexlab{a}}.
  Association for Computing Machinery.
\newblock ISBN 9781450349062.
\newblock \doi{10.1145/3123266.3123405}.
\newblock URL \url{https://doi.org/10.1145/3123266.3123405}.

\bibitem[Mao et~al.(2017{\natexlab{b}})Mao, Cheung, and She]{mao}
Hui Mao, Ming Cheung, and James She.
\newblock Deepart: Learning joint representations of visual arts.
\newblock In \emph{Proceedings of the 25th ACM International Conference on
  Multimedia}, MM '17, page 1183–1191, New York, NY, USA, 2017{\natexlab{b}}.
  Association for Computing Machinery.
\newblock ISBN 9781450349062.
\newblock \doi{10.1145/3123266.3123405}.
\newblock URL \url{https://doi.org/10.1145/3123266.3123405}.

\bibitem[Mensink and van Gemert(2014)]{Mensink2014}
Thomas Mensink and Jan van Gemert.
\newblock The rijksmuseum challenge.
\newblock In \emph{Proceedings of International Conference on Multimedia
  Retrieval}. {ACM}, April 2014.
\newblock \doi{10.1145/2578726.2578791}.
\newblock URL \url{https://doi.org/10.1145/2578726.2578791}.

\bibitem[Milani and Fraternali(2020)]{Milani_Fraternali_2020}
Federico Milani and Piero Fraternali.
\newblock A data set and a convolutional model for iconography classification
  in paintings.
\newblock \emph{arXiv:2010.11697 [cs]}, Nov 2020.
\newblock URL \url{http://arxiv.org/abs/2010.11697}.
\newblock arXiv: 2010.11697.

\bibitem[Misra et~al.(2016)Misra, Shrivastava, Gupta, and
  Hebert]{cross-stitch2016}
Ishan Misra, Abhinav Shrivastava, Abhinav Gupta, and Martial Hebert.
\newblock Cross-stitch networks for multi-task learning.
\newblock In \emph{CVPR}, pages 3994--4003. {IEEE} Computer Society, 2016.
\newblock \doi{10.1109/CVPR.2016.433}.

\bibitem[Pinkney(2020)]{pinkney2020ukiyoe}
Justin N.~M. Pinkney.
\newblock Aligned ukiyo-e faces dataset.
\newblock \url{https://www.justinpinkney.com/ukiyoe-dataset}, 2020.

\bibitem[Ridnik et~al.(2021)Ridnik, Baruch, Noy, and
  Zelnik{-}Manor]{DBLP:journals/corr/abs-2104-10972}
Tal Ridnik, Emanuel~Ben Baruch, Asaf Noy, and Lihi Zelnik{-}Manor.
\newblock Imagenet-21k pretraining for the masses.
\newblock \emph{CoRR}, abs/2104.10972, 2021.
\newblock URL \url{https://arxiv.org/abs/2104.10972}.

\bibitem[Russakovsky et~al.(2015)Russakovsky, Deng, Su, Krause, Satheesh, Ma,
  Huang, Karpathy, Khosla, and
  Bernstein]{Russakovsky_Deng_Su_Krause_Satheesh_Ma_Huang_Karpathy_Khosla_Bernstein_2015}
Olga Russakovsky, Jia Deng, Hao Su, Jonathan Krause, Sanjeev Satheesh, Sean Ma,
  Zhiheng Huang, Andrej Karpathy, Aditya Khosla, and Michael Bernstein.
\newblock Imagenet large scale visual recognition challenge.
\newblock \emph{International journal of computer vision}, 115\penalty0
  (3):\penalty0 211–252, 2015.

\bibitem[Sener and Koltun(2018)]{sener_neurips2018}
Ozan Sener and Vladlen Koltun.
\newblock Multi-task learning as multi-objective optimization.
\newblock In \emph{Proceedings of the 32nd International Conference on Neural
  Information Processing Systems}, NIPS'18, page 525–536, Red Hook, NY, USA,
  2018. Curran Associates Inc.

\bibitem[Simonyan et~al.(2014)Simonyan, Vedaldi, and
  Zisserman]{simonyan2014deep}
Karen Simonyan, Andrea Vedaldi, and Andrew Zisserman.
\newblock Deep inside convolutional networks: Visualising image classification
  models and saliency maps, 2014.

\bibitem[Strezoski and Worring(2017)]{strezoski2017omniart}
Gjorgji Strezoski and Marcel Worring.
\newblock Omniart: Multi-task deep learning for artistic data analysis.
\newblock \emph{arXiv preprint arXiv:1708.00684}, 2017.

\bibitem[Strezoski et~al.(2019)Strezoski, Noord, and
  Worring]{Strezoski_Noord_Worring_2019}
Gjorgji Strezoski, Nanne Noord, and Marcel Worring.
\newblock Many task learning with task routing.
\newblock In \emph{2019 IEEE/CVF International Conference on Computer Vision
  (ICCV)}, page 1375–1384. IEEE, Oct 2019.
\newblock ISBN 978-1-72814-803-8.
\newblock \doi{10.1109/ICCV.2019.00146}.
\newblock URL \url{https://ieeexplore.ieee.org/document/9010933/}.

\bibitem[Tan et~al.(2016)Tan, Chan, Aguirre, and Tanaka]{7533051}
Wei~Ren Tan, Chee~Seng Chan, Hernán~E. Aguirre, and Kiyoshi Tanaka.
\newblock Ceci n'est pas une pipe: A deep convolutional network for fine-art
  paintings classification.
\newblock In \emph{2016 IEEE International Conference on Image Processing
  (ICIP)}, pages 3703--3707, 2016.
\newblock \doi{10.1109/ICIP.2016.7533051}.

\bibitem[Tian et~al.(2020)Tian, Suzuki, Clanuwat, Bober-Irizar, Lamb, and
  Kitamoto]{tian2020kaokore}
Yingtao Tian, Chikahiko Suzuki, Tarin Clanuwat, Mikel Bober-Irizar, Alex Lamb,
  and Asanobu Kitamoto.
\newblock Kaokore: A pre-modern japanese art facial expression dataset, 2020.

\bibitem[Tian et~al.(2021)Tian, Clanuwat, Suzuki, and Kitamoto]{tian2021ukiyoe}
Yingtao Tian, Tarin Clanuwat, Chikahiko Suzuki, and Asanobu Kitamoto.
\newblock Ukiyo-e analysis and creativity with attribute and geometry
  annotation, 2021.

\bibitem[{van Noord} et~al.(2015){van Noord}, {Hendriks}, and
  {Postma}]{7123719}
N.~{van Noord}, E.~{Hendriks}, and E.~{Postma}.
\newblock Toward discovery of the artist's style: Learning to recognize artists
  by their artworks.
\newblock \emph{IEEE Signal Processing Magazine}, 32\penalty0 (4):\penalty0
  46--54, 2015.
\newblock \doi{10.1109/MSP.2015.2406955}.

\bibitem[Vandenhende et~al.(2020)Vandenhende, Georgoulis, and
  Van~Gool]{Vandenhende_Georgoulis_Van_Gool_2020}
Simon Vandenhende, Stamatios Georgoulis, and Luc Van~Gool.
\newblock Mti-net: Multi-scale task interaction networks for multi-task
  learning.
\newblock In Andrea Vedaldi, Horst Bischof, Thomas Brox, and Jan-Michael Frahm,
  editors, \emph{ECCV 2020}, pages 527--543. Springer International Publishing,
  2020.
\newblock \doi{10.1007/978-3-030-58548-8_31}.

\bibitem[Vernier et~al.(2020)Vernier, Caselles-Dupré, and
  Fautrel]{VERNIER2020100026}
Gauthier Vernier, Hugo Caselles-Dupré, and Peirre Fautrel.
\newblock Electric dreams of ukiyo: A series of japanese artworks created by an
  artificial intelligence.
\newblock \emph{Patterns}, 1\penalty0 (2):\penalty0 100026, 2020.
\newblock ISSN 2666-3899.
\newblock \doi{https://doi.org/10.1016/j.patter.2020.100026}.
\newblock URL
  \url{https://www.sciencedirect.com/science/article/pii/S266638992030026X}.

\bibitem[Westlake et~al.(2016)Westlake, Cai, and Hall]{Westlake2016}
Nicholas Westlake, Hongping Cai, and Peter Hall.
\newblock Detecting people in artwork with {CNNs}.
\newblock In \emph{Lecture Notes in Computer Science}, pages 825--841. Springer
  International Publishing, 2016.
\newblock \doi{10.1007/978-3-319-46604-0_57}.
\newblock URL \url{https://doi.org/10.1007/978-3-319-46604-0_57}.

\bibitem[Yang et~al.(2020)Yang, Quan, Nie, and
  Yang]{DBLP:journals/corr/abs-2012-14214}
Sen Yang, Zhibin Quan, Mu~Nie, and Wankou Yang.
\newblock Transpose: Towards explainable human pose estimation by transformer.
\newblock \emph{CoRR}, abs/2012.14214, 2020.
\newblock URL \url{https://arxiv.org/abs/2012.14214}.

\bibitem[Yin et~al.(2016)Yin, Monson, Honig, Daubechies, and Maggioni]{7472087}
Rujie Yin, Eric Monson, Elizabeth Honig, Ingrid Daubechies, and Mauro Maggioni.
\newblock Object recognition in art drawings: Transfer of a neural network.
\newblock In \emph{2016 IEEE International Conference on Acoustics, Speech and
  Signal Processing (ICASSP)}, pages 2299--2303, 2016.
\newblock \doi{10.1109/ICASSP.2016.7472087}.

\bibitem[Yosinski et~al.(2014)Yosinski, Clune, Bengio, and
  Lipson]{Yosinski_Clune_Bengio_Lipson_2014}
Jason Yosinski, Jeff Clune, Yoshua Bengio, and Hod Lipson.
\newblock How transferable are features in deep neural networks?
\newblock In Z.~Ghahramani, M.~Welling, C.~Cortes, N.~Lawrence, and K.~Q.
  Weinberger, editors, \emph{Advances in Neural Information Processing
  Systems}, volume~27. Curran Associates, Inc., 2014.
\newblock URL
  \url{https://proceedings.neurips.cc/paper/2014/file/375c71349b295fbe2dcdca9206f20a06-Paper.pdf}.

\bibitem[Zeiler and Fergus(2013)]{DBLP:journals/corr/ZeilerF13}
Matthew~D. Zeiler and Rob Fergus.
\newblock Visualizing and understanding convolutional networks.
\newblock \emph{CoRR}, abs/1311.2901, 2013.
\newblock URL \url{http://arxiv.org/abs/1311.2901}.

\bibitem[Zhang et~al.(2020)Zhang, Wang, Wu, Zhou, and Zhu]{9050545}
Quanshi Zhang, Xin Wang, Ying~Nian Wu, Huilin Zhou, and Song-Chun Zhu.
\newblock Interpretable cnns for object classification.
\newblock \emph{IEEE Transactions on Pattern Analysis and Machine
  Intelligence}, pages 1--1, 2020.
\newblock \doi{10.1109/TPAMI.2020.2982882}.

\end{thebibliography}
\end{document}